\documentclass[11pt]{article} 
\usepackage{rldm,palatino}
\usepackage{graphicx}
\usepackage{hyperref}
\usepackage{amsmath}
\usepackage{xspace}
\usepackage{float}
\usepackage{subcaption}

\def\Nat{{\rm I\kern\pIR N}}

\newcommand{\EE}[1]{\exptE\left[#1\right]}

\def\vec0{\mathbf 0}

\def\vecv{\mathbf v}
\def\vecw{\mathbf w}

\newcommand{\beq}{\begin{equation}}
\newcommand{\eeq}{\end{equation}}
\newcommand{\beqa}{\begin{eqnarray}}
\newcommand{\eeqa}{\end{eqnarray}}
\newcommand{\beqan}{\begin{eqnarray*}}
\newcommand{\eeqan}{\end{eqnarray*}}
\newcommand{\ben}{\begin{eqnarray*}}
\newcommand{\een}{\end{eqnarray*}}

\renewcommand{\EE}[2]{\mathbb{E}_{#1\!\!}\left[#2\right]}

\def\E#1{\EE{\,}{#1}}

\def\Nat{{\rm I\kern\pIR N}}

\def\vec0{{\bf{0}}}

\def\vecv{{\bf{v}}}
\def\vecw{{\bf{w}}}

\def\w{\vecw}

\def\vecw{{\bf{\bf w}}}

\def\vecv{{\bf{\bf v}}}

\def\ze{($\zeta$)\xspace}
\def\lab{($\lambda,\beta$)\xspace}
\def\forward{\texttt{forward}\xspace}
\def\turnaway{\texttt{turnaway}\xspace}
\usepackage{enumitem,amssymb}
\usepackage{pifont}
\newlist{todolist}{itemize}{10}
\setlist[todolist]{label=$\square$}

\usepackage{mleftright}\mleftright

\renewcommand{\EE}[2]{\mathbb{E}_{#1}\left[#2\right]}
\newcommand{\VV}[1]{\text{Var}\left[#1\right]}
\newcommand{\CV}[1]{\text{Cov}\left[#1\right]}
\newcommand{\given}[0]{\,\middle|\,}

\newcommand{\ex}[1]{\triangleright \, \text{#1}}
\renewcommand{\v}[1]{\bf{\bf{#1}}}

\usepackage[
backend=biber,
style=ieee,
]{biblatex}
\title{Importance Sampling Placement in Off-Policy Temporal-Difference Algorithms}

\author{
Eric Graves \\
Department of Computing Science \\
University of Alberta \\
\texttt{graves@ualberta.ca}
\And
Sina Ghiassian \\
Department of Computing Science \\
University of Alberta \\
\texttt{ghiassia@ualberta.ca} \\
}

\begin{document}

\allowdisplaybreaks

\maketitle

\begin{abstract}
A central challenge to applying many off-policy reinforcement learning algorithms to real world problems is the variance introduced by importance sampling.
In off-policy learning, the agent learns about a different policy than the one being executed.
To account for this difference, importance sampling ratios are often used, but can increase the variance of the algorithms and hence reduce the rate of learning.
Several variations of importance sampling have been proposed to reduce variance, with per-decision importance sampling being the most popular.
However, the update rules for most off-policy algorithms in the literature depart from per-decision importance sampling in a subtle way; they correct the entire TD error instead of just the TD target.
In this work, we show how this slight change can be interpreted as a control variate for the TD target, leading to reduced variance.
Experiments over a wide range of algorithms show this subtle modification results in improved performance.
\end{abstract}




\section{Off-Policy Temporal-Difference Algorithms}
Learning the long-term consequences of making decisions according to a specific policy is a central challenge in reinforcement learning.
This problem is known as the prediction problem, and involves an agent learning a value function for a given policy, often using some form of temporal-difference learning \parencite{sutton1988learning}.
One way of doing this is to simply execute the policy, observe the effects, and update an approximate value function---an approach known as \textit{on-policy} learning.
However, there are many applications where this approach can be expensive (e.g., advertising \parencite{bottou2013counterfactual}), dangerous (medicine \parencite{liao2021off}), or inefficient (robotics \parencite{smart2002effective} and educational applications \parencite{koedinger2013new}).

An alternative approach where the agent learns about a \textit{target policy} that is different from the \textit{behaviour policy} being executed is known as \textit{off-policy} learning.
Off-policy algorithms allow the agent to learn from experience generated by old policies (experience replay \parencite{lin1992self,schaul2015prioritized}), exploratory policies (Q-learning \parencite{watkins1992q}), human demonstrations \parencite{smart2002effective}, non-learning controllers, or even random behaviour.
They also enable offline learning for algorithms that are too computationally demanding to run online \parencite{levine2020offline}.
Perhaps most importantly for the development of AI, off-policy algorithms allow an agent to learn about many possible ways of behaving, in parallel, from a single stream of experience \parencite{sutton2011horde,white2015developing,klissarov2021flexible}.

There are several algorithms for learning value functions off-policy that have been proven to converge with function approximation, including the family of Gradient TD algorithms \parencite{sutton2009fast,hackman2013faster,ghiassian2020gradient}, Emphatic TD \parencite{sutton2016anemphatic}, Tree Backup \parencite{precup2000eligibility,touati2018convergent,ghiassian2022online}, Retrace \parencite{munos2016safe,touati2018convergent}, ABQ \parencite{mahmood2017multi}, and others \parencite{dann2014policy,white2016investigating,geist2014off,ghiassian2022online}.
Aside from Tree Backup and ABQ, all the aforementioned algorithms use importance sampling \parencite{rubinstein2016simulation} to 
correct for the difference in probability assigned to actions by the target and behaviour policies.
However, the variance introduced by importance sampling is a key limitation of these algorithms \parencite{precup2000eligibility}\parencite{liu2020understanding}.

A variety of importance sampling variants have been developed for off-policy learning in an attempt to reduce variance: per-decision \parencite{precup2000eligibility}, weighted \parencite{precup2000eligibility}, discounting-aware \parencite{sutton2018reinforcement}\parencite{mahmood2017incremental}, and stationary state distribution \parencite{hallak2017consistent}\parencite{liu2018breaking} importance sampling.
However, when inspecting the update rules for most algorithms that learn value functions off-policy, the placement of importance sampling ratios does not correspond to any of the known importance sampling variants.

In this extended abstract, we investigate this inconsistency and empirically compare the performance of several existing algorithms 
with versions that strictly implement per-decision importance sampling.
We find that the per-decision versions almost always perform worse, and show how 
scaling the entire TD error---as done by most existing algorithms---can be interpreted as a control variate, often reducing variance and improving performance.



\section{Importance Sampling Placement}
The goal of off-policy value function learning is to estimate the expected sum of future rewards (referred to as the \textit{value}) that would be received when executing target policy $\pi$ from each state, using observed rewards generated by executing behaviour policy $b$.
However, the behaviour policy may choose actions with different probabilities than the target policy.
To correct for this discrepancy, the observed sum of rewards can be scaled by the relative probability of taking each action in the trajectory under the target and behaviour policies, known as the importance sampling ratio and denoted $\rho_t = \pi(A_t|S_t) / b(A_t|S_t)$.
However, each reward only needs to be scaled by the importance sampling ratios that \textit{precede} it in the trajectory, as rewards cannot depend on decisions made in the future.
This is the idea behind the Per-Decision Importance Sampling-corrected return
\begin{align}
    G^{\text{PDIS}}_{t} &= \rho_{t} R_{t+1} + \gamma \rho_{t} \rho_{t+1} R_{t+2} + \gamma^2 \rho_t \rho_{t+1} \rho_{t+2} R_{t+3} + \ldots
    = \rho_t \left(R_{t+1} + \gamma G^{\text{PDIS}}_{t+1}\right) \label{eq:def_pdis_return}
\end{align}
whose expectation under the behaviour policy is equal to $v_\pi(S_t)$, and whose variance is often lower than scaling each reward by 
the importance sampling ratios for all actions in the trajectory, as is done in ordinary importance sampling \parencite{precup2000eligibility}.
The recursive nature of the PDIS return gives rise to an off-policy Bellman equation:
\begin{align}
    v_\pi(s) &= \EE{b}{G^{\text{PDIS}}_t \given S_t=s} \label{eq:def_vpi_pdis_return}
    \\
    &= \EE{b}{\rho_t \left(R_{t+1} + \gamma G^{\text{PDIS}}_{t+1}\right) \given S_t=s} &\ex{equation (\ref{eq:def_pdis_return})} \notag
    \\
    &= \EE{b}{\rho_t R_{t+1} \given S_t=s} + \gamma \EE{b}{\rho_t G^{\text{PDIS}}_{t+1} \given S_t=s} &\ex{linearity of expectation} \notag
    \\
    &= \EE{b}{\rho_t R_{t+1} \given S_t=s} + \gamma \EE{b}{\EE{b}{\rho_t G^{\text{PDIS}}_{t+1} \given S_{t+1}=s'} \given S_t=s} &\ex{law of total expectation} \notag
    \\
    &= \EE{b}{\rho_t R_{t+1} \given S_t=s} + \gamma \EE{b}{\rho_t \EE{b}{G^{\text{PDIS}}_{t+1} \given S_{t+1}=s'} \given S_t=s} &\ex{$\rho_t$ constant in inner expression} \notag
    \\
    &= \EE{b}{\rho_t R_{t+1} \given S_t=s} + \gamma \EE{b}{\rho_t v_\pi(S_{t+1}) \given S_t=s} &\ex{equation (\ref{eq:def_vpi_pdis_return})} \notag
    \\
    &= \EE{b}{\rho_t \left(R_{t+1} + \gamma v_\pi(S_{t+1})\right) \given S_t=s} &\ex{linearity of expectation} \notag
\end{align}
which yields an off-policy Bellman Error by subtracting $v_\pi(s)$ from both sides and replacing the true value function $v_\pi(s)$ with an approximate value function $\hat{v}_\pi(s,\vecw)$ parameterized by a weight vector $\vecw$:
\begin{align*}
    \text{BE}(s,\vecw) &= \EE{b}{\rho_t \left(R_{t+1} + \gamma \hat{v}_\pi(S_{t+1},\vecw)\right) \given S_t=s} - \hat{v}_\pi(s,\vecw)
\end{align*}
Samples of the Bellman Error are known as the Temporal-Difference error
and form the basis for off-policy temporal-difference learning algorithms that use importance sampling:
\begin{align}
    \delta_t &= \rho_t \left[R_{t+1} + \gamma \hat{v}_\pi(S_{t+1},\vecw)\right] - \hat{v}_\pi(s,\vecw) \label{eq:def_td_error_1}
\end{align}
However, most algorithm update rules in the literature use a different TD error that also multiplies $\hat{v}_\pi(s,\vecw)$ by $\rho_t$:
\begin{align}
    \Tilde{\delta_t} &= \rho_t \left[R_{t+1} + \gamma \hat{v}_\pi(S_{t+1},\vecw) - \hat{v}_\pi(s,\vecw)\right] \label{eq:def_td_error_2}
\end{align}
What is the rationale for scaling $\hat{v}_\pi(s,\vecw)$ by $\rho_t$?
After all, the value estimate for state $s$ does not need to be corrected with the importance sampling ratio for the action that follows it, and in fact the value estimate is not even a random variable! Scaling it is contrary to the conventional wisdom behind per-decision importance sampling; only the terms in the return that need to be corrected should be to avoid introducing variance.

It turns out that scaling $\hat{v}_\pi(s,\vecw)$ by $\rho_t$ can be interpreted as applying the method of control variates to the update target of the TD error $\left( \rho_t [R_{t+1} + \gamma \hat{v}_\pi(S_{t+1},\vecw)] \right)$.
A control variate is a random variable with a known expected value that is correlated with a random variable whose unknown expected value we seek to estimate.
Given an estimator $X$ with an unknown expected value, we can subtract the control variate $Y$ and add its known mean $\EE{}{Y}$ to obtain a new estimator with the same expected value, but with lower variance if $Y$ is correlated with $X$:
\begin{align*}
    \Tilde{\delta_t} &= \rho_t \left[R_{t+1} + \gamma \hat{v}_\pi(S_{t+1},\vecw) - \hat{v}_\pi(s,\vecw)\right] &\ex{equation (\ref{eq:def_td_error_2})}
    \\
    &= \rho_t \left[R_{t+1} + \gamma \hat{v}_\pi(S_{t+1},\vecw)\right] - \rho_t \hat{v}_\pi(s,\vecw) &\ex{distribute $\rho_t$}
    \\
    &= \rho_t \left[R_{t+1} + \gamma \hat{v}_\pi(S_{t+1},\vecw)\right] - \rho_t \hat{v}_\pi(s,\vecw) + \hat{v}_\pi(s,\vecw) - \hat{v}_\pi(s,\vecw) &\ex{add and subtract $\hat{v}_\pi(s,\vecw)$}
    \\
    &= \rho_t \left[R_{t+1} + \gamma \hat{v}_\pi(S_{t+1},\vecw)\right] - \hat{v}_\pi(s,\vecw) - \rho_t \hat{v}_\pi(s,\vecw) + \hat{v}_\pi(s,\vecw) &\ex{rearrange}
    \\
    &= \delta_t - \underbrace{\rho_t \hat{v}_\pi(s,\vecw)}_{Y} + \underbrace{\hat{v}_\pi(s,\vecw)}_{\E{Y}} &\ex{equation (\ref{eq:def_td_error_1})}
\end{align*}
Doing this does not introduce bias, as the expected value of $\rho_t \hat{v}_\pi(s,\vecw)$ is $\hat{v}_\pi(s,\vecw)$.
Analyzing the variance of $\Tilde{\delta_t}$ yields:
\begin{align*}
    \VV{\Tilde{\delta_t}} &= \VV{\delta_t} + \VV{\rho_t \hat{v}_\pi(s,\vecw)} - 2 \, \CV{\delta_t, \rho_t \hat{v}_\pi(s,\vecw)}
    \\
    &= \VV{\delta_t} + \VV{\rho_t \hat{v}_\pi(s,\vecw)} - 2 \, \CV{\rho_t \left(R_{t+1} + \gamma \hat{v}_\pi(S_{t+1},\vecw)\right), \rho_t \hat{v}_\pi(s,\vecw)}
\end{align*}
so we would expect the variance of $\Tilde{\delta_t}$ to be reduced relative to $\delta_t$ when there is a strong correlation between $\rho_t \left(R_{t+1} + \gamma \hat{v}_\pi(S_{t+1},\vecw)\right)$ and $\rho_t \hat{v}_\pi(s,\vecw)$.
As both are estimates of $v_\pi(s)$ and $\rho_t$ appears in both terms, it's very likely there exists a strong correlation between the two terms, especially when the value estimates are consistent.

\vspace{-.3cm}
\section{Experiment}

\begin{figure*}[t]
      \centering
      \includegraphics[width=0.6\linewidth]{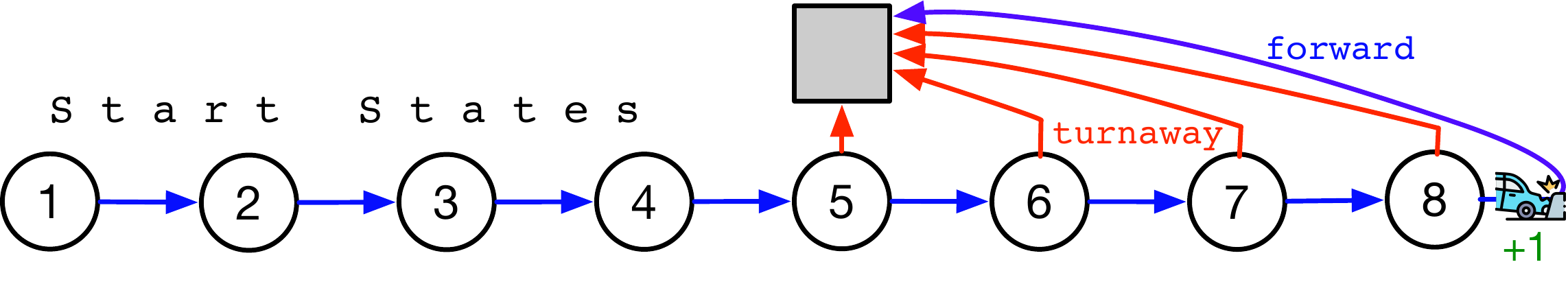}
      \caption{The Collision task. Episodes start in one of the first four states and end when the \forward action is taken from the eighth state, causing a crash and a reward of 1, or when the \turnaway action is taken in one of the last four states.}
      \label{fig:2_4-CollisionTask-Fig-v3}
\end{figure*}

To test this hypothesis, we conducted an experiment on the Collision task, a small environment with eight states and two actions shown in Figure \ref{fig:2_4-CollisionTask-Fig-v3} \parencite{ghiassian2021empirical}.
Under the target policy, the agent would always take the \forward action, while under the behavior policy it always takes the \forward action in the first four states and takes the \forward and \turnaway actions with equal probability in the four rightmost states.

We compared 2 versions of 10 different off-policy prediction learning algorithms, including GTD, GTD2, Proximal GTD2, HTD, Emphatic TD, Emphatic TD\lab, Off-policy TD, Vtrace, Tree Backup, and ABTD\ze.
One version used $\Tilde{\delta_t}$ (equation \ref{eq:def_td_error_2}) in the update rule (referred to as ``un-corrected $\v{w}^\top \v{x}$'' in Figure 2), and the other version used $\delta_t$ (equation \ref{eq:def_td_error_1}) in the update rule (referred to as ``corrected $\v{w}^\top \v{x}$'' in Figure 2).
We checked 19 values of the first step-size parameter, $\alpha$, for all algorithms: $\alpha = 2^{-x}$ where $x \in \{0, 1, 2, \ldots, 17, 18\}$.
For Gradient-TD algorithms we tried 15 values of $\eta$ where $\alpha_\vecv = \eta \times \alpha$.
The values of $\eta$ we checked were: $\eta = 2^{x}$ with $x \in \{-6, -5, \ldots, 7, 8\}$.
For Emphatic TD\lab we tried all combinations of the first step-size parameter $\alpha$ and $\beta\in\{0, 0.2, 0.4, 0.6, 0.8, 1\}$.
We set the bootstrapping parameter $\lambda$ to 0 for all algorithms that use it.\footnote{The bootstrapping parameter $\lambda$ interpolates between Temporal-Difference learning (biased, but lower variance) at $\lambda=0$ and Monte Carlo learning (unbiased, but extreme variance) at $\lambda=1$. We chose the lowest-variance setting of $\lambda$ because the variance of the algorithms can already be quite large due to importance sampling.}
In all experiments, we initialized the weight vector $\w_0=\vec0$ at the beginning of each run and ran the experiment for 20,000 time steps and 50 independent runs.
All the results presented are averages over the 50 runs and show the standard error over runs as a shaded region.

\begin{figure}[H]
    \centering
    \begin{subfigure}[b]{0.45\textwidth}
	    \includegraphics[width=\linewidth]{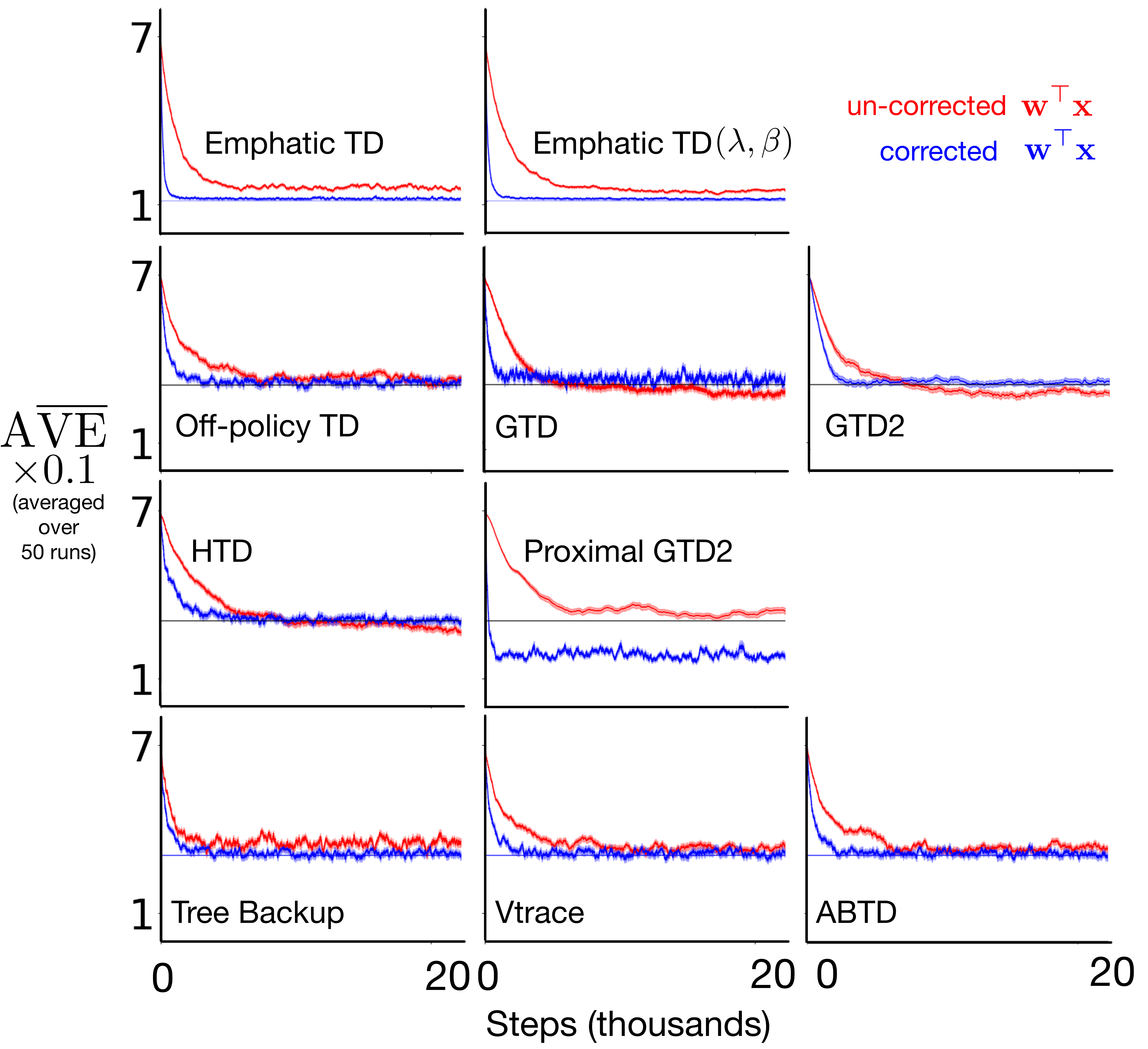}
	    \caption{Learning curves for each algorithm.}
	    \label{fig:learningcurverhoplacement}
    \end{subfigure}   
    \begin{subfigure}[b]{0.45\textwidth}
        \includegraphics[width=\linewidth]{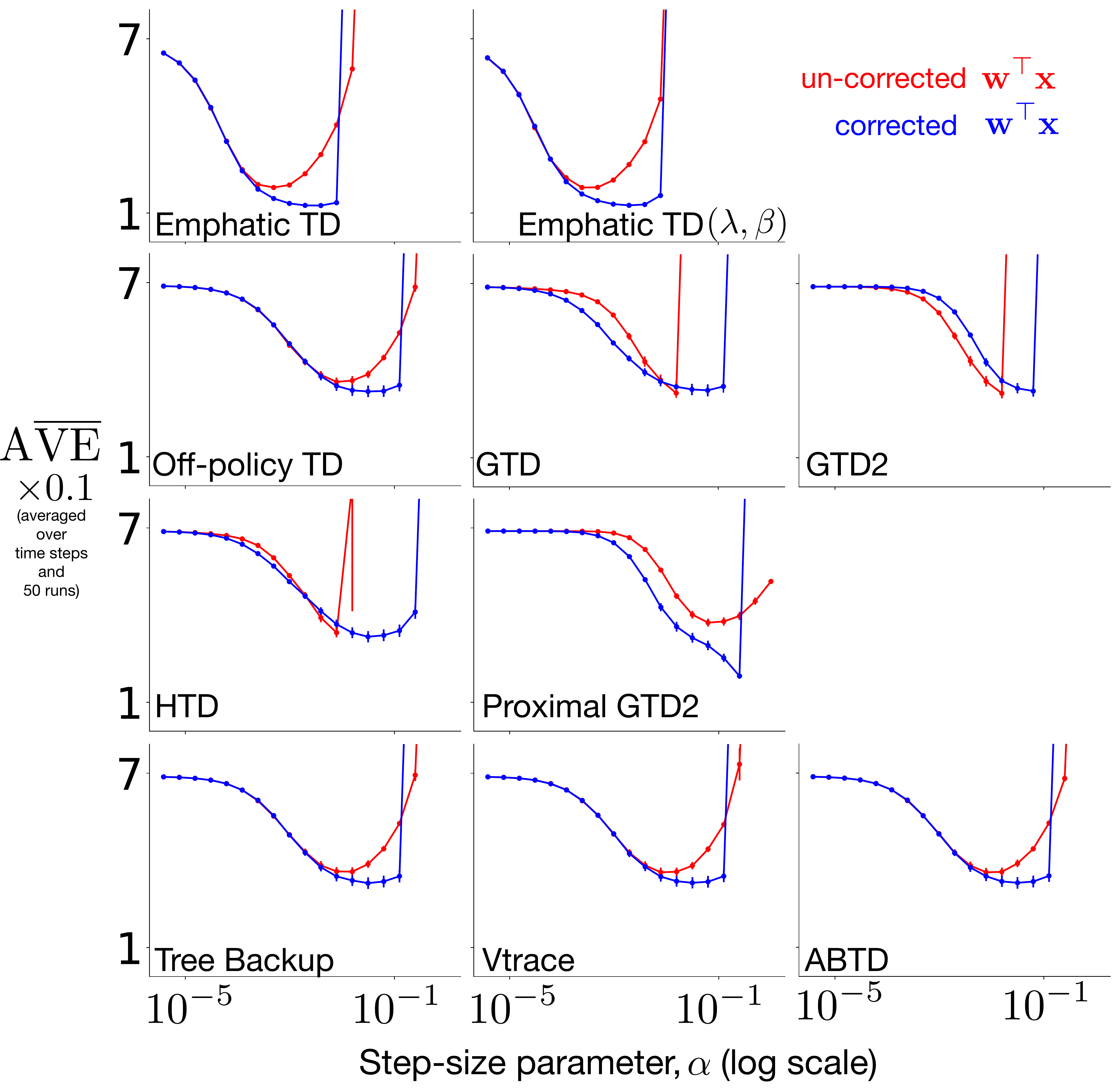}
        \caption{Sensitivity curves for each algorithm.}
        \label{fig:sensitivityrhoplacement}
    \end{subfigure}
    \caption{Learning curves and sensitivity curves for each algorithm on the Collision task. Blue is when the whole TD error term is corrected and red is when $v_\pi(s$) is not corrected.}
    \label{fig:my_label}
\end{figure}

The learning curves for the best algorithm instances (the parameter settings that resulted in the smallest area under the learning curve) for all algorithms are shown in Figure~\ref{fig:learningcurverhoplacement}.
We can see that in almost all cases, using $\Tilde{\delta_t}$ performed better than using $\delta_t$; the blue curve plateaued significantly sooner than the red curve, and often to a lower error.

The parameter sensitivity curves for all algorithms are shown in Figure~\ref{fig:sensitivityrhoplacement}. For algorithms that had more than one parameter, we plotted the sensitivity curve that included the best algorithm instance. We first found the algorithm instance that had the smallest area under the curve, and then fixed all the parameters, and plotted the results over the step-size parameter.
For all algorithms, when the whole TD error was corrected, the parameter sensitivity curve was wider, meaning that it is easier to choose a good step-size for the algorithm.

\vspace{-.3cm}
\section{Conclusion}
These results make it clear that correcting the whole TD error should be preferred over partially correcting the TD error when designing and implementing off-policy value function learning algorithms.
Correcting the whole TD error led to better performance for every algorithm involved, and also reduced every algorithm's sensitivity to the step-size parameter, making it easier to select good step-sizes.

\printbibliography

@book{sutton2018reinforcement,
  title={Reinforcement learning: An introduction},
  author={Sutton, Richard S and Barto, Andrew G},
  year={2018},
  publisher={MIT press}
}

@inproceedings{touati2018convergent,
  title={Convergent TREE BACKUP and RETRACE with function approximation},
  author={Touati, Ahmed and Bacon, Pierre-Luc and Precup, Doina and Vincent, Pascal},
  booktitle={International Conference on Machine Learning},
  pages={4955--4964},
  year={2018},
  organization={PMLR}
}

@article{klissarov2021flexible,
  title={Flexible Option Learning},
  author={Klissarov, Martin and Precup, Doina},
  journal={Advances in Neural Information Processing Systems},
  volume={34},
  year={2021}
}

@article{watkins1992q,
  title={Q-learning},
  author={Watkins, Christopher JCH and Dayan, Peter},
  journal={Machine learning},
  volume={8},
  number={3},
  pages={279--292},
  year={1992},
  publisher={Springer}
}

@phdthesis{white2015developing,
  title={Developing a predictive approach to knowledge},
  author={White, Adam},
  year={2015},
  school={University of Alberta}
}

@inproceedings{sutton2011horde,
  title={Horde: A Scalable Real--time Architecture For Learning Knowledge From Unsupervised Sensorimotor Interaction},
  author={Sutton, Richard S and Modayil, Joseph and Delp, Michael and Degris, Thomas and Pilarski, Patrick M and White, Adam and Precup, Doina},
  booktitle={Proceedings of the 10th International Conference on Autonomous Agents and MultiAgent Systems},
  year={2011}
}

@phdthesis{ghiassian2022online,
  title={Online Off-policy Prediction},
  author={Ghiassian, Sina},
  year={2022},
  school={University of Alberta}
}

@inproceedings{precup2000eligibility,
  title={Eligibility Traces for Off-Policy Policy Evaluation},
  author={Precup, Doina and Sutton, Richard S and Singh, Satinder P},
  booktitle={Proceedings of the Seventeenth International Conference on Machine Learning},
  pages={759--766},
  year={2000}
}

@article{bottou2013counterfactual,
  title={Counterfactual Reasoning and Learning Systems: The Example of Computational Advertising.},
  author={Bottou, L{\'e}on and Peters, Jonas and Qui{\~n}onero-Candela, Joaquin and Charles, Denis X and Chickering, D Max and Portugaly, Elon and Ray, Dipankar and Simard, Patrice and Snelson, Ed},
  journal={Journal of Machine Learning Research},
  volume={14},
  number={11},
  year={2013}
}

@article{lin1992self,
  title={Self-Improving Reactive Agents Based On Reinforcement Learning, Planning and Teaching},
  author={Lin, Long-Ji},
  journal={Machine Learning},
  volume={8},
  pages={293--321},
  year={1992}
}

@article{koedinger2013new,
  title={New potentials for data-driven intelligent tutoring system development and optimization},
  author={Koedinger, Kenneth R and Brunskill, Emma and Baker, Ryan SJd and McLaughlin, Elizabeth A and Stamper, John},
  journal={AI Magazine},
  volume={34},
  number={3},
  pages={27--41},
  year={2013}
}

@article{liao2021off,
  title={Off-policy estimation of long-term average outcomes with applications to mobile health},
  author={Liao, Peng and Klasnja, Predrag and Murphy, Susan},
  journal={Journal of the American Statistical Association},
  volume={116},
  number={533},
  pages={382--391},
  year={2021},
  publisher={Taylor \& Francis}
}

@article{levine2020offline,
  title={Offline reinforcement learning: Tutorial, review, and perspectives on open problems},
  author={Levine, Sergey and Kumar, Aviral and Tucker, George and Fu, Justin},
  journal={arXiv preprint arXiv:2005.01643},
  year={2020}
}

@inproceedings{schaul2015prioritized,
  title={Prioritized Experience Replay},
  author={Schaul, Tom and Quan, John and Antonoglou, Ioannis and Silver, David},
  booktitle={Proceedings of the 4th International Conference on Learning Representations},
  year={2016}
}

@inproceedings{smart2002effective,
  title={Effective reinforcement learning for mobile robots},
  author={Smart, William D and Kaelbling, L Pack},
  booktitle={Proceedings 2002 IEEE International Conference on Robotics and Automation (Cat. No. 02CH37292)},
  volume={4},
  pages={3404--3410},
  year={2002},
  organization={IEEE}
}

@inproceedings{sutton2009fast,
  title={Fast Gradient-Descent Methods for Temporal--Difference Learning with Linear Function Approximation},
  author={Sutton, Richard S and Maei, Hamid Reza and Precup, Doina and Bhatnagar, Shalabh and Silver, David and Szepesv{\'a}ri, Csaba and Wiewiora, Eric},
  booktitle={Proceedings of the 26th Annual International Conference on Machine Learning},
  year={2009}
}

@article{sutton2016anemphatic,
author = {Sutton, Richard S and Mahmood, A R and White, Martha},
title = {An Emphatic Approach to the Problem of Off--policy Temporal--Difference Learning},
journal = {The Journal of Machine Learning Research},
volume = {17},
year = {2016}
}

@inproceedings{ghiassian2020gradient,
  title={Gradient temporal-difference learning with regularized corrections},
  author={Ghiassian, Sina and Patterson, Andrew and Garg, Shivam and Gupta, Dhawal and White, Adam and White, Martha},
  booktitle={International Conference on Machine Learning},
  pages={3524--3534},
  year={2020},
  organization={PMLR}
}

@phdthesis{hackman2013faster,
  title={Faster Gradient-TD Algorithms},
  author={Hackman, Leah M},
  year={2013},
  school={University of Alberta}
}

@article{munos2016safe,
  title={Safe and efficient off-policy reinforcement learning},
  author={Munos, R{\'e}mi and Stepleton, Tom and Harutyunyan, Anna and Bellemare, Marc},
  journal={Advances in neural information processing systems},
  volume={29},
  year={2016}
}

@article{mahmood2017multi,
  title={Multi-step off-policy learning without importance sampling ratios},
  author={Mahmood, Ashique Rupam and Yu, Huizhen and Sutton, Richard S},
  journal={arXiv preprint arXiv:1702.03006},
  year={2017}
}

@article{dann2014policy,
  title={Policy evaluation with temporal differences: A survey and comparison},
  author={Dann, Christoph and Neumann, Gerhard and Peters, Jan and others},
  journal={Journal of Machine Learning Research},
  volume={15},
  pages={809--883},
  year={2014},
  publisher={Massachusetts Institute of Technology Press (MIT Press)/Microtome Publishing}
}

@inproceedings{white2016investigating,
  title={Investigating Practical Linear Temporal Difference Learning},
  author={White, Adam and White, Martha},
  booktitle={Proceedings of the 2016 International Conference on Autonomous Agents \& Multiagent Systems},
  pages={494--502},
  year={2016}
}

@article{geist2014off,
  title={Off-policy Learning With Eligibility Traces: A Survey},
  author={Geist, Matthieu and Scherrer, Bruno},
  journal={Journal of Machine Learning Research},
  volume={15},
  pages={289--333},
  year={2014}
}

@inproceedings{liu2020understanding,
  title={Understanding the curse of horizon in off-policy evaluation via conditional importance sampling},
  author={Liu, Yao and Bacon, Pierre-Luc and Brunskill, Emma},
  booktitle={International Conference on Machine Learning},
  pages={6184--6193},
  year={2020},
  organization={PMLR}
}

@phdthesis{mahmood2017incremental,
  title={Incremental Off-policy Reinforcement Learning Algorithms},
  author={Mahmood, Ashique},
  year={2017},
  school={University of Alberta}
}

@article{liu2018breaking,
  title={Breaking the curse of horizon: Infinite-horizon off-policy estimation},
  author={Liu, Qiang and Li, Lihong and Tang, Ziyang and Zhou, Dengyong},
  journal={Advances in Neural Information Processing Systems},
  volume={31},
  year={2018}
}

@inproceedings{hallak2017consistent,
  title={Consistent on-line off-policy evaluation},
  author={Hallak, Assaf and Mannor, Shie},
  booktitle={International Conference on Machine Learning},
  pages={1372--1383},
  year={2017},
  organization={PMLR}
}

@article{sutton1988learning,
  title={Learning to predict by the methods of temporal differences},
  author={Sutton, Richard S},
  journal={Machine learning},
  volume={3},
  number={1},
  pages={9--44},
  year={1988},
  publisher={Springer}
}

@book{rubinstein2016simulation,
	title={Simulation and the Monte Carlo method},
	author={Rubinstein, Reuven Y and Kroese, Dirk P},
	volume={10},
	year={2016},
	publisher={John Wiley \& Sons}
}

@article{ghiassian2021empirical,
  title={An Empirical Comparison of Off-policy Prediction Learning Algorithms on the Collision Task},
  author={Ghiassian, Sina and Sutton, Richard S},
  journal={arXiv preprint arXiv:2106.00922},
  year={2021}
}

\end{document}